# Robotic Electrospinning Actuated by Non-Circular Joint Continuum Manipulator for Endoluminal Therapy*

Zicong Wu, Chuqian Lou, Zhu Jin, Shaoping Huang, *Student Member*, *IEEE*, Ning Liu, Yun Zou, Mirko Kovac, Anzhu Gao*, *Member*, *IEEE*, Guang-Zhong Yang*, *Fellow, IEEE*

*Abstract—* Electrospinning has exhibited excellent benefits to treat the trauma for tissue engineering due to its produced micro/nano fibrous structure. It can effectively adhere to the tissue surface for long-term continuous therapy. This paper develops a robotic electrospinning platform for endoluminal therapy. The platform consists of a continuum manipulator, the electrospinning device, and the actuation unit. The continuum manipulator has two bending sections to facilitate the steering of the tip needle for a controllable spinning direction. Non-circular joint profile is carefully designed to enable a constant length of the centreline of a continuum manipulator for stable fluid transmission inside it. Experiments are performed on a bronchus phantom, and the steering ability and bending limitation in each direction are also investigated. The endoluminal electrospinning is also fulfilled by a trajectory following and points targeting experiments. The effective adhesive area of the produced fibre is also illustrated. The proposed robotic electrospinning shows its feasibility to precisely spread more therapeutic drug to construct fibrous structure for potential endoluminal treatments.

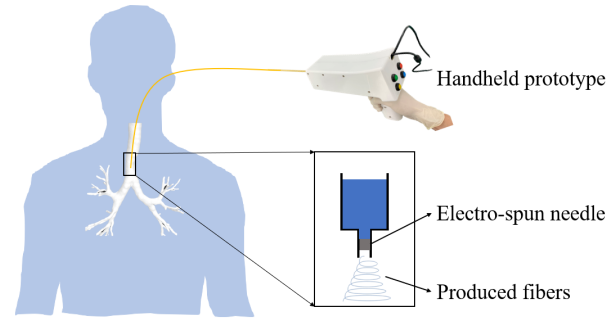

Fig. 1. A diagram of robotic electrospinning for endoluminal therapy.

## I. INTRODUCTION

Compared to traditional open surgeries, minimally invasive surgery has been gradually recognized as the gold standard to minimize the invasiveness during surgical procedures. It can reduce trauma, shorten recovery time, and decrease post-operative complications significantly [1]. Experienced a far way and progressed much in accuracy and dexterity, nowadays the minimally invasive surgeries are always evolving with the integration of endoscope technology, inserting surgical instruments through orifices of the human body to perform the therapeutic operations in the means of the endoluminal intervention [2].

In recent decades, remarkable advances of robots are gradually illustrating their tremendous potential for surgical applications [3]. Its significant impact on addressing the challenges in applying minimally invasive operations or endoluminal interventions into clinical scenarios has also reshaped traditional surgical procedures. The assistance from surgical robots can greatly enhance the safety and efficiency of the surgeons' operation, with reduced operative time, recovery period, blood loss, and contamination [4]. Benefited from their remarkable flexibility and locomotion capability, the continuum manipulators and soft robots are desired candidates that can be adapted to endoluminal interventions among all kinds of robots, which enables a safer manipulation [5]. Conventionally, the miniaturized continuum manipulators are preferred to help deliver and steer the surgical instruments into the deep site of the human body through natural or artificial orifices [6,7].

Recently, the conception of tissue engineering has gained popularity broadly, due to its potential to be applied as a repair or reconstruction technique for surgical operations, especially in reconstructive surgeries. It has exhibited great effectiveness in nerve repair [8], bone reconstruction [9], vascular ligature and suture [10], all of which are common and frequent operations during surgical operations. In the other hand, electrospinning has been developed as an innovative fabrication technique to produce micro/nano fibres using a variety of drug or materials. Its attractive feature of producing polymer fibres with a wide range of diameters reveals its great potential of broad applications. Specifically, significant popularities have been gained from medical area, including tissue engineering [11], drug delivery [12], dural repair and reconstruction [13], and organic hemostasis [14]. When fulfilling electrospinning, the polymer solution is always pressurized to flow through the metal needle where the voltage is applied. Then continuous ultra-fine fibres can be produced

Research supported in part by the National Natural Science Foundation of China (62003209); the SJTU Global Strategic Partnership Fund (2019 SJTU-CUHK), China; the Foundation of National Facility for Translational Medicine (Shanghai) (TMSK-2020-106), China; the State Key Laboratory of Robotics and Systems (HIT) (SKLRS-2020-KF-18); the Open Project Fund from Shenzhen Institute of Artificial Intelligence and Robotics for Society, China (AC01202005012). (*Corresponding authors: Anzhu Gao and Guang-Zhong Yang*)

Zicong Wu, Shaoping Huang, and Yun Zou are with the Institute of Medical Robotics and Department of Bioengineering, Shanghai Jiao Tong University, 200240, Shanghai, P. R. China.

Chuqian Lou is with Imperial College London and Empa-Swiss Federal Laboratories for Materials Science and Technology.

Ning Liu is with Precision Robotics (Hong Kong) Limited, Hong Kong, 999077, China.

Mirko Kovac is with Aerial Robotics Laboratory, Imperial College London. London SW7 2AZ, United Kingdom.

Anzhu Gao is with the Institute of Medical Robotics and Department of Automation, Shanghai Jiao Tong University, and the Key Laboratory of System Control and Information Processing, Ministry of Education, Shanghai 200240, China; Shanghai Engineering Research Center of Intelligent Control and Management, Shanghai 200240, China.

Guang-Zhong Yang is with the Institute of Medical Robotics, Shanghai Jiao Tong University, 200240, Shanghai, P. R. China.

with a diameter from 3 nm to more than 5 um, which can be decided by the setup [15].

To perform electrospinning for hemostasis during the endoluminal interventions, the capability of precise targeting is required. However, most of existed surgical applications of electrospinning are designed to be operated manually, where undesired operation maybe introduced accidentally by inexpertness, physical fatigue, or intrinsic tremble [16]. Air flow-based targeting has been employed to perform the electrospinning, but this may cause contamination or adhesive to the surrounding normal tissues and non-target areas [15]. To address the limitations, continuum manipulators can be ideal candidates to perform precise targeting of electrospinning direction to produce fibrous structures toward targeted areas.

In our previous work, a miniaturized laser-profiled thin-walled continuum manipulator with interlocked circular joints has been developed to help steer the imaging probe to access the desired targets in the distal airways [17-20]. It can be well adapted to perform the electrospinning in the endoluminal environment, whose central lumen will be designed to place through a polymer fluid tube where polymer solution is properly pressurized and fed. However, the centreline length of the continuum manipulator always changes during deflection, which would inevitably cause the translational motion of the tube placed in the inner lumen of the continuum manipulator. Such translational motion could affect the accurate fluid transmission from the proximal end to the distal end. Therefore, it is necessary to develop a large-lumen continuum manipulator that enables a constant length of the centreline during deflection to perform tissue engineering.

This paper develops a robotic electrospinning platform for endoluminal therapy using a continuum manipulator with non-circular joint contour. Fig. 1 shows a diagram of robotic electrospinning using the developed continuum manipulator and electrospinning device. The developed robotic platform can successfully generate micro/nano fibrous structure to target surfaces, as well as enhance the steerability of the distal electrospinning jet with better orientation capability. It shows great potential to mix multiple targeted drugs into the solution to produce fibrous structure, thereby accomplishing the long-term targeted therapy.

This paper is organized as follows: Section II introduces the proposed robotic electrospinning platform that is actuated by a continuum manipulator with novel non-circular joint, and the design of continuum manipulator, joint profile, as well as electrospinning device are detailed in each subsection separately; Section III describes how this proposed robotic electrospinning platform validated by experiments and the acquired validation result is exhibited in each part; especially the capability of maintaining the constant length of centreline and producing fibres with micro/nano diameters.

## II. ROBOTIC ELECTROSPINNING PLATFORM

The robotic electrospinning platform developed by us consists of a continuum manipulator, the electrospinning equipment and a microflow injection pump, as shown in Fig. 2. The employed continuum manipulator enables flexible steering in the very confined environment, which benefits guaranteeing the safety of the patients as well as achieving obstacle avoidance efficiently. This is beneficial for surgical operations during the endoluminal intervention, especially when it comes to the operations on the delicate and fragile anatomical tissues. A fluid tube is placed through the central lumen of the continuum manipulator to feed the solution to perform electrospinning and produce fibres. The capabilities of flexible steering, remarkable locomotion, and maintaining a constant length of the continuum manipulator's centreline will facilitate electrospinning for endoluminal therapy.

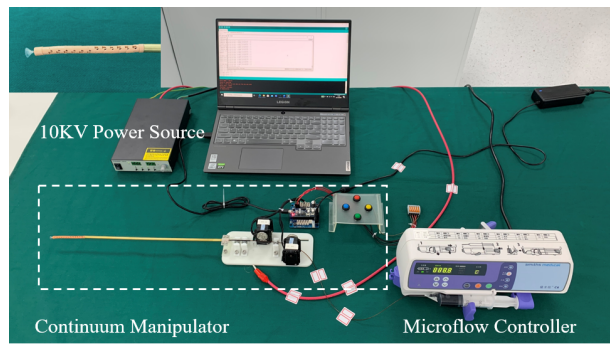

Fig. 2. A preliminary platform of robotic electrospinning. It consists of a continuum manipulator, a high voltage power source, and a microflow injection controller.

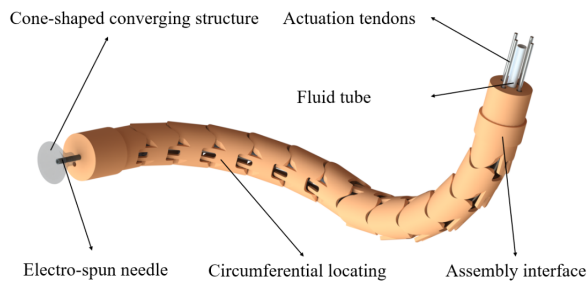

Fig. 3. The structure of the developed continuum manipulator with two bending sections. A metal conductive needle is mounted at its tip to inject produced fibers under high voltage, and a cone structure is employed to restrain and converge the direction of injected fibers.

TABLE I. DESIGN REQUIREMENTS OF THE CONTINUUM ROBOT WITH NONCIRCULAR JOINTS

| Items | Requirement |
| --- | --- |
| Outer diameter | 5 mm |
| Diameter of central lumen | ≥ 1.2 mm |
| Wall thickness | ≥ 1.5 mm |
| The length of robot | ≥ 50 mm |
| Deflection | Bidirectional |
| Manufacturing | 3D Printing |
| No. of segments | 13 |
| Material | RC 90 Resin |

### A. Design of Continuum Manipulator with Two Bending Sections

Given the required length of the continuum manipulator, the number of segments can be roughly decided. The other important parameter determining the continuum manipulator's morphology is the number of sections. Traditionally, more sections are preferred to enhance flexibility and locomotion capabilities. Such design can always accomplish better steering capability because each section can be actuated independently. In this paper, our continuum manipulator is designed to enable the independent steering of two individual sections. A detailed design specification of the continuum manipulator is shown in Table I. The central lumen has a diameter that is no less than 1.2 mm to allow the fluid tube to pass through and perform the desired electrospinning during endoluminal intervention.

Our continuum manipulator is fabricated by 3D printing using P4K printer (EnvisionTEC, USA) with RC90 material, with a total length of 69.3 mm. It consists of 13 segments and can be divided into two individual sections that are connected by the connecting segment, as Fig. 3 shows. These segments can be noted as $S_i$ ($i$ = 0, 1, 2, …, 12) from the base of the continuum manipulator to its distal tip. Segments $S_0$ to $S_5$ form the first individual section while $S_7$ to $S_{12}$ constitute the second discrete section, with segment $S_6$ connecting two sections. Such design is specifically meaningful in the confined operating environment, like the bronchotracheal or the rectum. Higher degree of freedoms is always desired and critical under these surgical scenarios. The discrete sections enable the steering in two different planes, which are perpendicular with each other.

For each section, two tendons are used to actuate the steering in two different directions and the bending direction is decided by which tendon is pulled. The wheel plate is designed and assembled onto each MX 28 motor (Dynamixel, South Korea) where actuation tendons are connected at the designated positions. Pulleys are used to keep the tension of these actuation tendons. When the motor is rotating in clockwise direction, one tendon will be pulled while the other one will be released so that this section can be steered in a specific direction. Similarly, when the motor is rotating in anti-clockwise direction the section would be deflected in the opposite direction. Moreover, the angle of deflection can be decided by the pulled length of actuation tendons. In each segment, a disc with one large central lumen and four circumferentially arranged small lumens are designed to place the plastic fluid tube and the actuation tendons through the robot body, respectively. Specifically, the first discrete section can be steered toward upward or downward direction, while the second individual section enables the bending toward left and right directions.

The MX 28 motors are controlled by an OpenCM 9.04 C board (Dynamixel, South Korea). An OpenCM 485 exp board (Dynamixel, South Korea) is employed to support the power supply and communication. A 12V direct current power source is provided by an adapter to power the motors. The Arduino sketch is uploaded onto the OpenCM 9.04 C controller board and the position information can be acquired from the serial monitor of the Arduino IDE in real-time. Four buttons are used to control the steering of the first section toward upward and downward direction, and the steering of the second one toward left and right directions, respectively. Each rotational step is set as 0.4 degrees and the corresponding length of actuation tendons is around 6 mm.

*B. Design of Non-circular Joint Profile*

This section demonstrates how the design is performed to acquire the joint profile for our continuum manipulator to enable the constant length of its centreline regardless of the deflection angle. To determine the profile of the desired joint, a mathematical model should be formulated to describe the geometrical relationship between any two adjacent segments and their joint within the continuum manipulator. The constant length of the centreline at an arbitrary angle during deflection is applied as the boundary conditions, then the equations to describe the contour profile guaranteeing this function can be solved and further optimized applying other constraints.

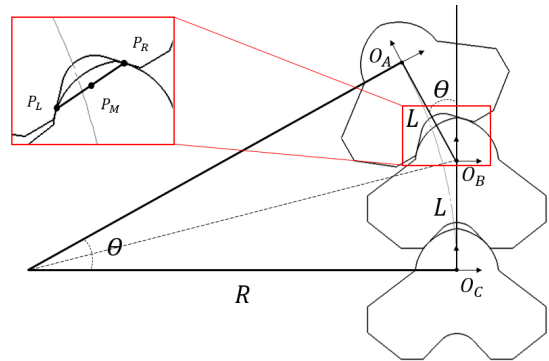

Fig. 4. The geometrical relationship between two adjacent segments within the continuum robot. The distance between the origin point of their coordinate frame is $L$, and the deflection angle is labelled as $\theta$.

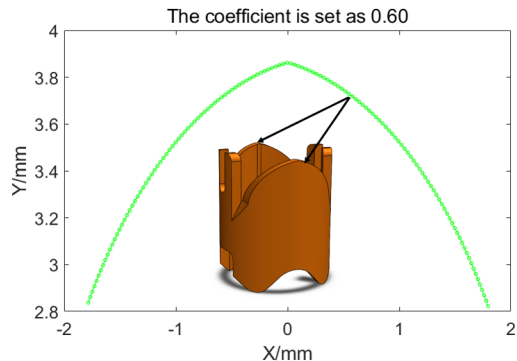

Fig. 5. Generated joint profile based on theoretical calculation and the developed segment model using the determined contour.

We have previously developed continuum manipulators to perform the delivery of imaging probe for optical biopsies. However, our laser-profiled continuum manipulator are circular joints, whose centreline length will change obviously during deflection. When performing the electrospinning using the continuum manipulator, a polyimide tube is inserted through the central lumen of it to feed the solution to the electron-spun needle that is placed at the tip of the continuum manipulator. Thus, the relative translational motion of the fluid tube within the central lumen may result in the inference with other inner placed components. Moreover, such motions can cause the inflation or deflation of the tube and the fluid within it, which may alter even intermit the feeding of electrospinning solution accidentally. The fluid tube may even be struck or broken, resulting in contaminate or infection during the endoluminal intervention, threatening the lives of patients.

Fig. 4 demonstrates three adjacent segments and the joint between them in detail. For each segment, a coordinate is attached on, which are noted as the coordinates $O_A$, $O_B$, and $O_C$ from top to bottom. The origins of these coordinates are $A$, $B$, and $C$, respectively. The distance between adjacent coordinate origins, i.e., $L_{AB}$ and $L_{BC}$ equals to $L$. The backbone of the continuum manipulator is rigid. Thus, the centreline length of the continuum manipulator is primarily affected by joint profile. By analysing two adjacent segments, the entire length of the centreline can be estimated. A mathematics model can be proposed to derive the relative translational shift of the centreline.

When the continuum manipulator keeps straight the bending angle $\theta$ equals to zero, which could be regarded as the initial state and the length of the centreline $L_{AC}$ is $2L$. Assuming coordinate $C$ as the base point, the tip position, which is the coordinate origin of the upper segment $A$, is (0, $2L$). When the continuum manipulator is bending, it can be deflected into a curve, as is a section of arc of a circle. If segment $A$ is bending with an angle of $\theta$, the radius $R$ and the arc length $\widehat{AB}$ can be determined as (1) and (2). The coordinate of the segment $(x, y)$, can also be derived as (3) and (4), respectively. Obviously, the calculated arc length in (2) is always less than the original length of the central lumen $2L$ under bending condition, which means that the optical fibre will be translated with regards to the disk along the axial direction. Thus, the inner placed optical fibre would shift out of the continuum manipulator.

$$R = \frac{L}{2\tan\left(\frac{\theta}{2}\right)} \quad (1)$$

$$\widehat{AB} = \frac{L\theta}{2\tan\left(\frac{\theta}{2}\right)} \quad (2)$$

$$x = \frac{L}{2\tan\left(\frac{\theta}{2}\right)}(\cos(\theta) - 1) \quad (3)$$

$$y = \frac{L}{2\tan\left(\frac{\theta}{2}\right)}\sin(\theta) \quad (4)$$

To derive the non-circular joint profile, it is necessary to meet the conditions to ensure the constant centreline length of the coordinate origins between two adjacent segments, i.e., $2L$ at arbitrary deflected angle $\theta$. As shown in Fig. 4, the contact points between adjacent segments $A$ and $B$ are $P_L$ and $P_R$, respectively. It could be observed that only the line connecting two contact points $P_L$ and $P_R$ is always perpendicular with the deflected centreline and divide the centreline into two segments with equal length, the constant length of the centreline can be guaranteed. When the deflected centreline from coordinate origins of segments $A$ and $B$, that is the arc length equals to the original length of $2L$, the radius of the circle $R^*$ can be determined as (5). The midpoint of the arc $P_m$, which is also the midpoint of the connecting line of contact points, can be calculated as (6). The length of the connecting line is denoted as $S$. To facilitate determining the contour, the related parameters can be normalized by a coefficient $N$ in (7). This coefficient generally decides the dimension of the segment. Consequently, the coordinates of the contact points can be calculated as $P_L$ and $P_R$, shown as (8) and (9), separately.

$$R^* = \frac{2L}{\theta} \quad (5)$$

$$P_m = \left(\frac{2L}{\theta}\cos\left(\frac{\theta}{2}\right) - \frac{2L}{\theta}, \frac{2L}{\theta}\sin\left(\frac{\theta}{2}\right)\right) \quad (6)$$

$$S = N * L \quad (7)$$

$$P_L = \left(\frac{2L}{\theta}\cos\left(\frac{\theta}{2}\right) - \frac{2L}{\theta} - \frac{S}{2}\cos\left(\frac{\theta}{2}\right), \frac{2L}{\theta}\sin\left(\frac{\theta}{2}\right) - \frac{S}{2}\sin\left(\frac{\theta}{2}\right)\right) \quad (8)$$

$$P_R = \left(\frac{2L}{\theta}\cos\left(\frac{\theta}{2}\right) - \frac{2L}{\theta} + \frac{S}{2}\cos\left(\frac{\theta}{2}\right), \frac{2L}{\theta}\sin\left(\frac{\theta}{2}\right) + \frac{S}{2}\sin\left(\frac{\theta}{2}\right)\right) \quad (9)$$

The determined coordinates of contact points form the trajectory thus the contour profile, as shown in Fig. 5. To find the desired joint profile, the contour has to be intersected and close entirely to avoid the rotation obstruction or overlapping. Thus, the optimized contour has been determined by finding the critical condition of an entirely closed contour, with the normalization factor $N$ set as 0.60. The value of $L$ is selected as 3.5 mm here, and the corresponding distance between two contact points is 2.1 mm. This noncircular contour guarantees that the line connecting two contact point can always be divided into two equal sections by the arc, no matter what the deflected angle is. At each joint, a planar rotation of $\pm\frac{\pi}{4}$ is guaranteed without any interference or overlapping. This can be verified by model and prototype.

Assuming the original coordinates of the contact points is $(x_c, y_c)$ when the continuum manipulator keeps straight, both contact points will experience the same rotational motion and translational displacement. Thus, a transformation matrix can be derived to describe the coordinates with regards to the deflective angle of the upper segment within any two adjacent segments of the continuum root, shown as (10). Anticlockwise is specified as the positive rotational direction here.

$$\begin{bmatrix} X_{deflected} \\ Y_{deflected} \\ 1 \end{bmatrix} = \begin{bmatrix} \cos\left(\frac{\theta}{2}\right) & -\sin\left(\frac{\theta}{2}\right) & \frac{2L}{\theta}\cos\left(\frac{\theta}{2}\right) - \frac{2L}{\theta} \\ \sin\left(\frac{\theta}{2}\right) & \cos\left(\frac{\theta}{2}\right) & \frac{2L}{\theta}\sin\left(\frac{\theta}{2}\right) \\ 0 & 0 & 1 \end{bmatrix} \begin{bmatrix} x_c \\ y_c \\ 1 \end{bmatrix} \quad (10)$$

### C. Design of Electrospinning Device

The electrospinning device consists of the high-voltage power source, a syringe, and an electro-spun needle. The high voltage is necessary to produce fibres. It is supplied to the needle tip custom made by the super-fine sliver wire with a diameter of 7 microns that is placed in the central lumen along with the fluid tube. The super-fine silver wire is inserted into a tube to reduce the potential influence from the produced heat on the surroundings. The PVP bio-compatible polymer is chosen for this application for its easy formation of fibres at laboratory conditions. The PVP powders (Boai Inc., China) are dissolved in ethyl alcohol with 15 wt%. The optimum spinning parameter for this solution is 10 kV voltage supply, 0.5mL/h feeding speed with a spun distance of around 12 cm.

In this robotic electrospinning platform developed by us, the electro-spun needle with a length of 3 mm and a diameter of 0.55 mm is mounted at the continuum manipulator's tip by a cap-like interface. A super-fine silver wire with a diameter of 7 microns and a fluid tube with a diameter of 1.12 mm are placed into the large central lumen and through the whole continuum manipulator. After coming out of the continuum manipulator body, the silver wire, actuation tendons, and the fluid tube are placed into a brass tube with a length of 30 cm, to offer the continuum manipulator abilities to access the deep and hard-to-reach sites of the human body.

An interface is designed at the distal end of the brass tube, to facilitate the arrangement of actuation tendons, the fluid tube as well as the conductive wire. All their rest part coming out of the brass tube are protected by a soft tube to prevent potential damage, with a section of a heat-shrinkable tube connecting the soft tube and the designed interface. The end of fluid tube is also connected to a syringe by the heat-shrinkable tube to prevent leakage of electrospinning solution.

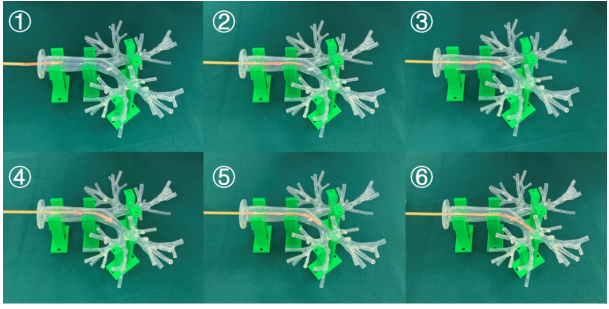

Fig. 6. The phantom study of simulating the endoluminal intervention using the developed robotic platform to validate its steering and locomotion capability in the confined environment. Sequences 1 to 6 exhibit the intervention procedure from inserting into the major bronchus to entering deep site of the sub-bronchus.

The syringe is assembled with a microflow injection pump to make the PVP solution be feed at proper rate and pressure.

## III. Experimental Results and Discussion

To validate the performance of our proposed robotic electrospinning platform, a series of experiments have been performed. In this section, how the validation experiments are carried out is described in detail, including validating the continuum manipulator's steering capacity, the capability of maintaining a constant length of its centreline, as well as the effectiveness of electrospinning. The experimental results are acquired and analysed in each sub-section, respectively. Discussions are also provided in this part.

### A. Validation of Steering Capacity

Firstly, the steering capacity of the continuum manipulator is validated by phantom study. An intervention procedure into the transparent bronchus phantom made from silica gel (Trandomed, China) is employed. Benefited by its remarkable flexibility, the continuum manipulator can change morphology to keep aligned with airways' centreline, without any collision with the surroundings after inserting into the major bronchus.

The intervention procedure is well presented by Fig. 6. The design of a two-section structure with segment connector facilitates its motion in the very confined environment, especially when entering the sub-bronchus. It illustrates the great clinical significance of independent steering of multiple sections. In this experimental study, the translational motion of the continuum manipulator is adjusted by pushing manually, which can be further improved using a step motor to perform the forward and backward movement.

To determine the steering range of the continuum manipulator, a series of deflection angles in each direction are tested for these sections, as shown in Fig. 7. These segments are fabricated using resin, thus only limited pulling force is applied in case of structural failure. To reduce the friction force exerted among these segments, the PMX-200 silicone fluid with a viscosity of 1000 CPS (Xiameter Inc., German) is used to lubricate these components to avoid the structural damage that might be introduced accidentally.

### B. Validation of Constant Length of Centreline

To validate the robot's capability of maintaining a constant length of its centreline, a plastic tube with a diameter of 1.1 mm is inserted into the central lumen of the continuum manipulator and keep aligned with the manipulator's centreline. The tube's diameter is a bit smaller than the diameter of the central lumen to try to keep their centrelines aligned at the meantime guarantee that the tube can be translationally moved. One of its ends is aligned with the proximal end of the robot and fixed with Loctite 4013 medical glue. The other end is also keep aligned with the distal end of the continuum manipulator but it's free thus be able to move out of the robot body translationally. Afterward, different deflection angles are performed and a series of measurements of the tube's length coming out of the robot's tip is acquired by a vernier caliper correspondingly.

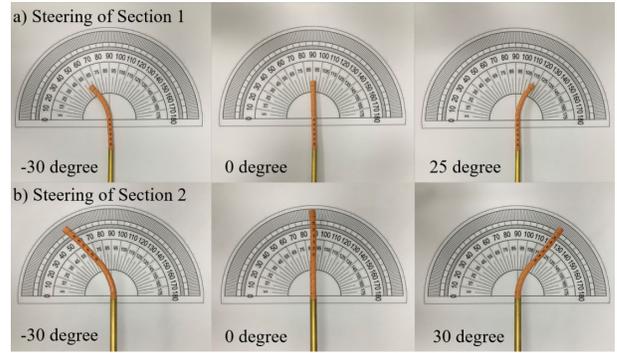

Fig. 7. The steering range is validated. The bending angles toward left, right, upward and downward directions are all around 30 degrees.

In this proposed continuum manipulator, all the joints are fabricated with the same contour and two sections are designed to achieve independent steering. Thus, only the section near the robot's distal tip is tested in the deflection experiments. At each deflection angle, three times of measurements have been performed to calculate the average value in order to reduce the random errors. It is observed that within the deflection range of $\pm 30$ degrees, the length change of the centreline didn't exceed 0.1 mm, where rotation toward right is set as the positive direction. This matches the theoretical calculation in previous section and it proves that the developed non-circular contour indeed plays the function of compensation for the orginal shorten centreline successfully. The observed slight change of centreline can be explained by the manufacturing error, imperfect assembly, and measurement errors. To further validate the capability of maintaining a constant length of the continuum manipulator's centreline, the EM sensor can be placed at the tip of the continuum manipulator to acquire more accurate position information.

### C. Validation of Steerable Electrospinning

After validating the continuum manipulator's performance, this developed platform is employed to fulfill electrospinning. The electrospinning solution is made by PVP powder with the absolute alcohol as the solvent. It is provided by an off-the-shelf syringe with a capacity of 30 ml to produce fibres under high voltage and proper pressure. The microflow injection pump is set as 0.5 ml/hour and the applied voltage is set as 10 kV, which are the referenced feeding speed and voltage for electrospinning using this solution. A cone-shaped structure is
designed to limit the deposition area of the fibres to avoid the adhesive to the surroundings and contamination.

On a piece of A4 paper in green background, an area in elliptical shape is labelled by black line as the targeted area, on which the produced fibres are deposited and turn into a thin polymer film. To improve the effect, a thin metal board is placed behind the paper and the ground voltage is applied on them. Thus, the produced fibres can be gathered and deposited on the paper due to the electrostatic field instead of diffusing without order and cause adhesion all over around. To support the imaging using scanning electron microscope, the other sample is made, where the spun fibres are deposited on the tin film. Then, the microcosmic structure of the deposited fibres can be acquired by the MIRA3 FEG-SEM (TESCAN *Inc.*, Czech Republic) to validate the feasibility and effectiveness.

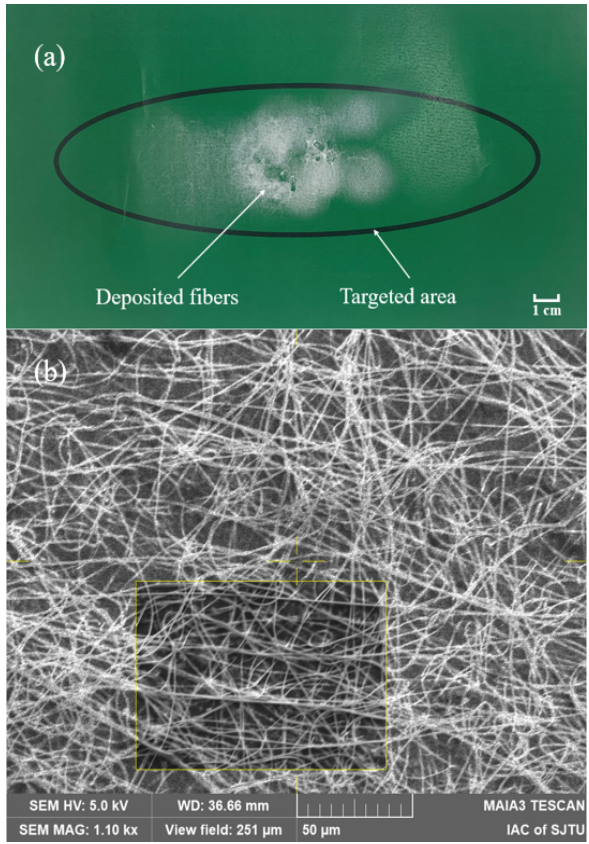

Fig. 8. Validation of robotic electrospinning for fibre forming. (a) The deposited film of fibre using electrospinning with the robot. (b) The microcosmic structure of deposited fibres under scanning electron microscope.

Under 10 kV voltage, the fibres are produced from the electro-spun needle that is placed at the tip of the continuum manipulator and then deposited at the A4 paper placed at the distance of 12 cm ahead of the continuum manipulator's tip. The continuum manipulator is controlled to spin in the labelled area. Fig. 8(a) shows the deposited fibres in the targeted area on the paper. It can be observed that the produced fibres are deposited and turned into a thin film in white colour. The electrospinning is extremely sensitive to the working environment, especially the electrostatic field. Thus, it is difficult to make the produced fibres distributed uniformly. Besides, the surrounding s will affect the electrostatic field and make it hard to maintain the injection direction of the fibres. Discontinuity and inhomogeneity phenomenon can be observed in the film in Fig. 8(a), which can be explained by the sudden disruption of electrospinning caused by imperfect grounding or operators' destabilization.

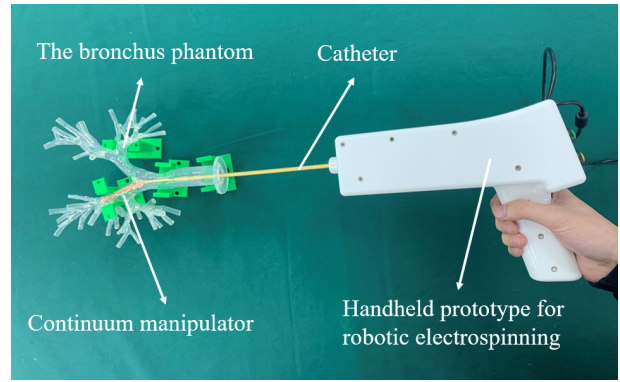

Fig. 9. The developed prototype of electrospinning platform actuated by a continuum manipulator for endoluminal intervention on a bronchus phantom.

During experiments, it is observed that multiple fibres are generated simultaneously at the tip of the electro-spun needle and manipulated by the electrostatic field. The injection directions are easy to be disrupted by the random fluctuations of the field. This explains why some fibres are deposited out of the targeted area and always leads to adhesion to the surrounding tissues. Thus, maintaining a good grounding for the target and stable working environment is important to perform the desired electrospinning. Fig. 8(b) illustrates the microcosmic structure of the produced fibres under the scanning electron microscope. The produced fibres exhibit homogeneity in diameters at submicron level, which can be deposited into films for the endoluminal therapies of the targeted tissues. The experimental result illustrates the feasibilities and effectiveness of the proposed robotic electrospinning platform. A handheld prototype has also been developed to facilitate the operation, as shown in Fig. 9. The function of force sensing with multiple degree of freedom can be integrated at the tip of the continuum manipulator to maintain a desired contact with targeted tissue and further guarantee the patient's safety [21-23]. More precise and advanced control scheme can also be investigated to improve its performance [24].

## IV. CONCLUSION

This paper develops a robotic electrospinning platform for endoluminal therapies. A novel continuum manipulator with two bending sections has been developed to achieve the robotic electrospinning. The non-circular joint profile has been designed to maintain a constant length of the continuum manipulator's centreline during deflection. This avoids the potential relative motion between the manipulator inner lumen and the fluid tube located inside it, therefore reducing the effect to the fluid transmission. Besides, the developed continuum manipulator based on our proposed non-circular joint profile exhibits good steering capability in the confined environment. It has been integrated with the electrospinning device and mimicked the procedure of endoluminal hemostasis successfully. The fibres at the micron scale produced and deposited at the desired areas to generate thin polymer films. The integration of the continuum manipulator and electrospinning illustrates great potential in clinical scenarios for steerable targeted therapies with patient's specific anatomy and drug medicines.